\documentclass[letterpaper]{article} 
\usepackage[]{aaai25}  
\usepackage{times}  
\usepackage{helvet}  
\usepackage{courier}  
\usepackage[hyphens]{url}  
\usepackage{graphicx} 
\usepackage{natbib}  
\usepackage{caption} 
\usepackage{algorithm}
\usepackage{algorithmic}
\usepackage{booktabs}
\usepackage{amsfonts}       
\usepackage{nicefrac}       
\usepackage{microtype}      
\usepackage{xcolor}         
\usepackage{amsmath}
\usepackage{dsfont}
\usepackage{amssymb}
\usepackage{subfigure}
\usepackage{multirow}
\usepackage{makecell}
\usepackage{bbding}

\usepackage{hyperref}
\hypersetup{
	colorlinks=true,
	linkcolor=blue,
	filecolor=blue,      
	citecolor=teal,
}
\usepackage{newfloat}
\usepackage{listings}
\DeclareCaptionStyle{ruled}{labelfont=normalfont,labelsep=colon,strut=off} 
\floatstyle{ruled}
\newfloat{listing}{tb}{lst}{}
\floatname{listing}{Listing}
\title{Bridging the Gap for Test-Time Multimodal Sentiment Analysis}
\author{
    Zirun Guo\equalcontrib,
    Tao Jin\thanks{Corresponding author},
    Wenlong Xu\equalcontrib,
    Wang Lin,
    Yangyang Wu
}
\affiliations{
    Zhejiang University\\
    zrguo.cs@gmail.com
}

\begin{document}

\maketitle

\begin{abstract}
    Multimodal sentiment analysis (MSA) is an emerging research topic that aims to understand and recognize human sentiment or emotions through multiple modalities. However, in real-world dynamic scenarios, the distribution of target data is always changing and different from the source data used to train the model, which leads to performance degradation. Common adaptation methods usually need source data, which could pose privacy issues or storage overheads. Therefore, test-time adaptation (TTA) methods are introduced to improve the performance of the model at inference time. Existing TTA methods are always based on probabilistic models and unimodal learning, and thus can not be applied to MSA which is often considered as a multimodal regression task. In this paper, we propose two strategies: \textbf{C}ontrastive \textbf{A}daptation and \textbf{S}table \textbf{P}seudo-label generation (CASP) for test-time adaptation for multimodal sentiment analysis. The two strategies deal with the distribution shifts for MSA by enforcing consistency and minimizing empirical risk, respectively. Extensive experiments show that CASP brings significant and consistent improvements to the performance of the model across various distribution shift settings and with different backbones, demonstrating its effectiveness and versatility.
\end{abstract}

\begin{links}
\link{Code}{https://github.com/zrguo/CASP}
\end{links}

\section{Introduction}
Multimodal Sentiment Analysis (MSA) aims to understand and interpret human sentiment or emotions expressed through multiple modalities such as text, video and audio. Compared to traditional multimodal sentiment analysis which focuses on analyzing text data to determine the sentiment or emotion associated with a particular piece of text, MSA combines information from various modalities to gain a deeper understanding of sentiment. With the success of multimodal learning, MSA has attracted much attention~\cite{7742221, tsai-etal-2019-multimodal, guo2024multimodal}. However, in real-world dynamic scenarios, the test data distribution is always changing which could lead to performance degradation of the model. For example, the model is trained on a multimodal sentiment analysis dataset in English but tested on a Chinese dataset, or the model is trained on a dataset with no background noise in the audio modality but tested on a dataset with a lot of background noise in the audio modality. Besides, in the video modality, different people have different facial traits. All of these things can be regarded as distribution shifts during the test stage and could lead to performance degradation of the model.

\begin{figure}
  \centering
  \includegraphics[width=0.85\linewidth]{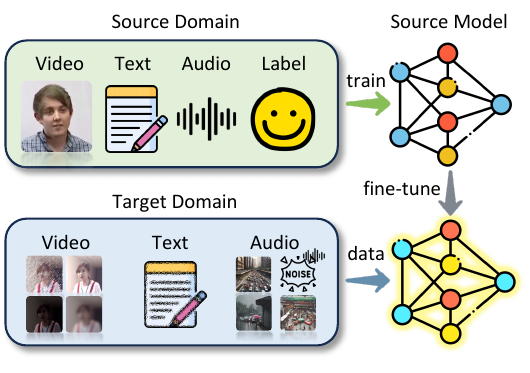}
  \caption{Test-time adaptation for multimodal sentiment analysis. The source domain data is used for source model training and is unavailable during the adaptation process. The target domain data is unlabeled.}
  \label{setting}
  \vskip -0.1in
\end{figure}

To address distribution shifts at test time, Test-Time Adaptation (TTA) is proposed~\cite{wang2020tent}. TTA aims at overcoming the distribution gaps between source and target domains during test time without accessing the source data and the labels of the target data~\cite{wang2020tent}. Figure~\ref{setting} presents the setting of TTA for MSA. However, existing methods can not be applied to MSA for two main reasons. On the one hand, MSA is often regarded as a regression task~\citep{bagher-zadeh-etal-2018-multimodal, tsai-etal-2019-multimodal} where the label is a score representing the intensity of the sentiment. However, most existing methods~\cite{wang2020tent, chen2022contrastive, wang2022continual, zhang2022memo} are based on probabilistic models for classification. For example, \citet{wang2020tent} propose entropy minimization for adaptation which is a function for classification tasks and can not be applied to regression tasks. On the other hand, MSA is a multimodal task whereas existing methods mainly focus on unimodal tasks, overlooking the property of multimodal data and thus can not be applied to MSA. For instance, \citet{zhang2022memo} perform different data augmentations on images to estimate marginal output distribution averaged over augmentations. However, it is hard to implement data augmentations on multimodal data, especially on extracted features instead of raw data.

Based on the above observations, in this paper, we propose \textbf{C}ontrastive \textbf{A}daptation and \textbf{S}table \textbf{P}seudo-label generation (CASP) for test-time adaptation for multimodal sentiment analysis. Specifically, our adaptation process has two stages: i) we introduce a contrastive adaptation strategy via modality random dropout to enforce consistency and improve the generalization ability of the model, meanwhile generating pseudo labels every few epochs and ii) we calculate the average value of the difference between the pseudo labels generated in stage one to select high-confident pseudo labels for self-training. The two stages deal with the TTA problem for MSA from two perspectives. Concretely, the contrastive adaptation strategy adapts the model by consistency regularization while the self-training with stable pseudo labels adapts the model by minimizing the empirical risk.

We conduct extensive experiments on three multimodal sentiment analysis datasets: CMU-MOSI~\cite{7742221}, CMU-MOSEI~\cite{bagher-zadeh-etal-2018-multimodal} and CH-SIMS~\cite{yu-etal-2020-ch}. We use different backbones to validate CASP's universality. The results show that CASP outperforms all the baselines significantly and consistently, demonstrating its superiority and versatility. Then, ablation experiments are conducted to measure the contribution of contrastive adaptation and stable pseudo labels and for a better understanding of CASP. To summarize, our contributions are as follows:
\begin{itemize}
  \item We propose test-time adaptation techniques CASP for multimodal sentiment analysis to alleviate the distribution shifts between the source domain and target domain data. To the best of our knowledge, CASP is the first TTA method for \textit{multimodal regression tasks}.
  \item We propose two novel strategies to address the distribution shifts of the target domain: contrastive adaptation to enforce consistency and stable pseudo-label generation to minimize the empirical risk.
  \item We show that CASP brings significant and consistent performance improvements to TTA for MSA across a range of settings and different backbones. The experimental results demonstrate the superiority and versatility of CASP.
\end{itemize}

\section{Related Work}
\textbf{Multimodal Sentiment Analysis.} 
Multimodal Sentiment Analysis (MSA) aims to predict sentiment intensity using multiple modalities such as text, video and audio. The main challenge of MSA is how to integrate information from different modalities effectively. Currently, there are mainly two types of fusion strategies: feature-level fusion (early fusion) and decision-level fusion (late fusion). Feature-level fusion methods~\cite{lazaridou2015combining, liang2018multimodal, wang2019words} combine the features extracted from different modalities to create a unified feature representation via concatenation or other methods before feeding it into the network. Different from feature-level methods, decision-level methods~\cite{tsai-etal-2019-multimodal, yu-etal-2020-ch} process different modalities separately and integrate them into the final decision. MISA~\citep{10.1145/3394171.3413678} projects each modality to two distinct subspaces to provide a holistic view of the multimodal data. MMIM~\citep{han2021improving} hierarchically maximizes the mutual information in unimodal input pairs and between multimodal fusion result and unimodal input to maintain task-related information. UniMSE~\citep{hu-etal-2022-unimse} proposes a knowledge-sharing framework that unifies MSA and MER to improve the performance. However, all these methods assume that the train and test data come from the same distribution. When there is a distribution shift between the train and test data, the performance of these models will degrade.

\begin{figure*}
  \centering
  \includegraphics[width=0.83\linewidth]{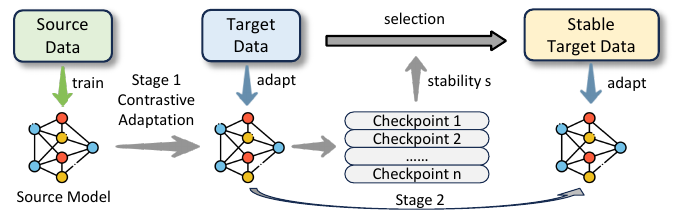}
  \caption{The overall framework of CASP. The adaptation process of CASP has two stages. Stage 1: contrastive adaptation to enforce consistency via modality random dropout. Stage 2: utilizing the checkpoints generated in Stage 1 to select high-confident pseudo labels for self-training. The two stages address the distribution shifts by consistency regularization and empirical risk minimization respectively.}
  \label{overall}
\end{figure*}

\noindent\textbf{Test-time Adaptation.} 
Test-Time Adaptation (TTA) refers to the adaptation of a pre-trained model to new target domain data without having access to the source domain data and the labels of target domain data. Unlike domain adaptation which requires access to both source and target data for adaptation, test-time adaptation methods do not require any data from the source domain and any label from the target domains. Among the various categories, one notable category is online TTA methods~\cite{wang2020tent,liang2020we,zhang2022memo}. For example, TENT~\cite{wang2020tent}, the first TTA approach, takes a pre-trained model and adapts it to the test data by updating the trainable parameters in normalization layers using entropy minimization. Source Hypothesis Transfer (SHOT)~\cite{liang2020we} proposes to update only the encoder parameters and align source and target representation by entropy minimization and pseudo-labeling. Recently, some works have delved into the multimodal test-time adaptation~\cite{shin2022mm, yang2023test}. MM-TTA~\cite{shin2022mm} proposes two complementary modules within and across the modalities to obtain reliable pseudo labels. READ~\cite{yang2023test} proposes reliable fusion against reliability bias and a novel objective function for robust multi-modal adaptation. In addition to online TTA methods, another category is robust TTA methods~\cite{niu2022towards, zhou2023ods}. This kind of method takes some challenging issues into account such as single sample and label shifts. Recently, some researchers have started exploring continual TTA methods~\cite{wang2022continual, gan2023decorate, wang2024continual} which deal with the continually changing domain shifts in real-world scenarios.

However, the methods mentioned above all consider the probabilistic model for classification tasks. Therefore, they can not be applied to regression tasks such as sentiment analysis and image quality assessment. One recent work~\cite{roy2023test} proposes auxiliary tasks to enable TTA for regression tasks. However, it is a unimodal framework that needs an image augmentation strategy, which can not be applied to multimodal tasks. In contrast, this paper proposes a TTA method for multimodal regression tasks.

\section{Methodology}\label{s3}
In this section, we will introduce our proposed method CASP. First, we will formulate our problem setting in Section~\ref{s31}. Then, we will introduce a contrastive adaptation strategy via modality random dropout in Section~\ref{s32}. Finally, in Section~\ref{s33}, we will introduce a stable pseudo-label generation strategy. The overall framework of our method is presented in Figure~\ref{overall}.

\subsection{Problem Formulation}\label{s31}
In MSA tasks, there are usually three modalities: audio, video and text. Therefore, we define the source domain data as $\mathcal S=\{(\boldsymbol{s}_i, y_i)\}_{i=1}^{N_s}$ where $N_s$ is the number of data and $\boldsymbol{s}_i=(s_i^a, s_i^v, s_i^t)$ represents the audio, video and text modality, respectively. In the TTA setting, we first pre-train our model on the source domain data $\mathcal S$. Suppose our model consists of the encoder $\mathcal M$ to get the feature representations and the prediction head $\mathcal F$ to get the final predictions, the output of the model is:
\begin{equation}
  \hat y = \mathcal F_{\theta_f}(\mathcal M_{\theta_m}(\boldsymbol{s}_i)),\quad \boldsymbol{s}_i\in\mathcal S
\end{equation}
where $\theta_m$ and $\theta_f$ are the parameters of $\mathcal M$ and $\mathcal F$, respectively. In MSA tasks, the loss function is often L1 loss~\cite{tsai-etal-2019-multimodal, yu-etal-2020-ch, guo2024multimodal}. Therefore, the optimization process is:
\begin{equation}
  \theta_m^*, \theta_f^*=\mathop{\arg\min}\limits_{\theta_m, \theta_f} |\hat y - y|
\end{equation}

Then we discard the source domain data and use the target domain data for adaptation. We define target domain data as $\mathcal T=\{\boldsymbol{x}_i\}_{i=1}^{N_t}$ where $N_t$ is the number of data. The labels of the target domain are unavailable.

\subsection{Contrastive Adaptation}\label{s32}
\begin{figure}
  \centering
  \includegraphics[width=\linewidth]{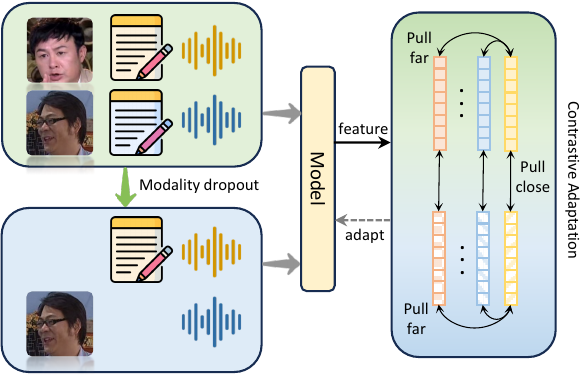}
  \caption{The overview of contrastive adaptation strategy. We randomly drop a modality to generate new data. Then we enforce the representations of the original data and the new data closer and distance the representation of the original data from the other representations in the batch.}
  \label{con}
\end{figure}

During the adaptation process, the predictions of the model are expected to be consistent when different kinds of data augmentation strategies are implemented. Some existing TTA methods~\cite{zhang2022memo,wang2022continual} impose data augmentation strategies and calculate the average probability distribution of the augmented data to minimize the entropy or as pseudo labels. However, in multimodal regression tasks, we can neither calculate probability distributions nor perform data augmentation. Since many multimodal tasks use extracted features instead of raw data~\cite{tsai-etal-2019-multimodal, yu-etal-2020-ch, guo2024multimodal}, data augmentation becomes difficult to implement. In order to enforce consistency, we introduce a contrastive adaptation strategy via modality random dropout to improve the generalization ability of the model. The overall diagram of the contrastive adaptation strategy is presented in Figure~\ref{con}.

Specifically, given $\boldsymbol{x}_i \in \mathcal T$, we impose a random dropout of modalities and replace the missing modality with $\boldsymbol{0}$ or other fixed value. We denote the data after modality random dropout as $\boldsymbol{x}_i^{\text{aug}}$, where different missing modality cases can be considered as different data augmentation strategies. We can obtain the feature representations $\boldsymbol{h}$ of $\boldsymbol{x}_i$ and $\boldsymbol{x}_i^{\text{aug}}$ following:
\begin{equation}
  \boldsymbol{h}_i=\mathcal M (\boldsymbol{x}_i),\quad \boldsymbol{h}_i^{\text{aug}}=\mathcal M (\boldsymbol{x}_i^{\text{aug}})
\end{equation}

To impose consistency regularization, we want to bring $\boldsymbol{h}_i$ and $\boldsymbol{h}_i^{\text{aug}}$ closer together and move $\boldsymbol{h}_i$ and other representations in the batch further apart. Therefore, we consider a modified \textit{NT-Xent} loss~\cite{chen2020simple}. Let $\textrm{sim}(\boldsymbol{u},\boldsymbol{v})=\boldsymbol{u}^\top\boldsymbol{v}/||\boldsymbol{u}||||\boldsymbol{v}||$ denote the dot product between $\ell_2$ normalized $\boldsymbol{u}$ and $\boldsymbol{v}$ (\textit{i.e.} cosine similarity). The loss function for a positive example $(\boldsymbol{h}_i, \boldsymbol{h}_i^{\text{aug}})$ is as follows:
\begin{equation}
  \ell_{\boldsymbol{h}_i, \boldsymbol{h}_i^{\text{aug}}}=-\log\frac{\exp(\textrm{sim}(\boldsymbol{h}_i, \boldsymbol{h}_i^{\text{aug}})/\tau)}{\sum_{k=1}^K\mathds 1_{[k\neq i]}\exp(\textrm{sim}(\boldsymbol{h}_i, \boldsymbol{h}_k)/\tau)}
\end{equation}
where $K$ is the batch size, $\tau$ is a temperature parameter, and $\mathds 1_{[k\neq i]}\in \{0,1\}$ is a sign function evaluating to 1 if $k\neq i$. The total loss can be written as:
\begin{equation}
  \mathcal L = \frac{1}{2K}\sum_{k=1}^K (\ell_{\boldsymbol{h}_k, \boldsymbol{h}_k^{\text{aug}}} + \ell_{\boldsymbol{h}_k^{\text{aug}}, \boldsymbol{h}_k})
\end{equation}

For stability and efficiency, previous works~\cite{wang2020tent, roy2023test} usually reconcile the distribution shifts by updating normalization layers. In our method, we follow these works, only updating the linear and lower-dimensional feature modulation parameters (\textit{i.e.} normalization layers). Through contrastive adaptation strategy, the model will learn more generalizable features and become more consistent.

\subsection{Stable Pseudo-label Generation}\label{s33}
Some previous TTA methods~\cite{wang2022continual, zhang2022memo} propose to generate high-confident pseudo labels for entropy minimization. Due to the probabilistic models, it is easy to generate pseudo labels to measure confidence (\textit{i.e.} the probability of each class). However, the output of a regression model is non-probabilistic, thus making it hard to measure confidence. In this subsection, we propose to measure the confidence of pseudo labels via a dynamic training process. Concretely, in Section~\ref{s32}, we introduce a contrastive adaptation strategy to improve the generalization ability of the model and make the model more consistent. Suppose the model is adapted for $E$ epochs, we denote the model of epoch $e$ as $F_e=(\mathcal M_{\theta_m^e}, \mathcal F_{\theta_f^e})$ where $e=1,2,\cdots,E$. Then the predictions of the model $F_e$ can be expressed as:
\begin{equation}
  \hat y_e = F_e(\boldsymbol{x})
\end{equation}
where $\hat y_e$ represents the predictions of all the training samples using $F_e$. For every epoch $e$, we can calculate the pseudo labels $\hat y_e$. During this dynamic process, some of the pseudo labels change a lot while some of them remain stable. Based on this observation, we propose to select those stable pseudo labels as high-confident pseudo labels and those that change a lot as low-confident pseudo labels. For efficiency and high quality, we calculate the pseudo labels of all the training samples every $M$ epochs. Mathematically, we calculate the difference between every two consecutive checkpoints, and use the average difference to measure the stability $s$:
\begin{equation}
  s = \frac{1}{\lfloor\frac{E}{M}\rfloor}\sum_{i=0}^{\lfloor\frac{E}{M}\rfloor-1}|\hat y_{iM}-\hat y_{(i+1)M}|
\end{equation}
where $\lfloor\frac{E}{M}\rfloor$ represents the largest integer not exceeding $E/M$ and $\hat{y}_0$ denotes the labels generated by the source model. After obtaining the stability $s$, we set a threshold to select high-confident pseudo labels. Specifically, we use $\lambda$-quantiles as the threshold. When $s$ is smaller than the threshold, we select the sample as our self-training sample. When $s$ is larger than the threshold, we discard the sample. For the values of pseudo labels, we use the average value of $\hat y$ of all the checkpoints:
\begin{equation}
  \tilde{y} = \frac{1}{\lfloor\frac{E}{M}\rfloor+1}\sum_{i=0}^{\lfloor\frac{E}{M}\rfloor}\hat{y}_{iM}
\end{equation}

Then, we can obtain the self-training dataset $\mathcal T_{\text{train}}=\{\boldsymbol{x}_i, \tilde{y}_i\}_{i=1}^{N_{\text{train}}}$ where $N_{\text{train}}$ is the number of selected high-confident samples. Then we use $\mathcal T_{\text{train}}$ for training.

\begin{table*}[]
  \centering
  \resizebox{\textwidth}{!}{%
  \begin{tabular}{@{}c|l|ccc|ccc|ccc|ccc|ccc@{}}
  \toprule
  \multirow{2}{*}{Backbone}     & \multirow{2}{*}{Method} & \multicolumn{3}{c}{MOSEI$\rightarrow$SIMS} & \multicolumn{3}{c}{MOSI$\rightarrow$SIMS} & \multicolumn{3}{c}{MOSI$\rightarrow$MOSEI} & \multicolumn{3}{c}{SIMS$\rightarrow$MOSI} & \multicolumn{3}{c}{SIMS$\rightarrow$MOSEI} \\ \cmidrule(l){3-17}
  &&ACC&F1&MAE&ACC&F1&MAE&ACC&F1&MAE&ACC&F1& MAE&ACC&F1&MAE\\ \midrule
  \multirow{6}{*}{\makecell[c]{Late\\Fusion}}  
  & Source&60.96&63.09&2.01&39.17&39.12&2.10&66.57&67.42&1.25&40.12&45.46&2.18&47.14&57.47&1.77\\
  & ST&62.01&65.19&\underline{1.95}&40.48&39.55&2.05&\underline{67.41}&\underline{67.90}&\underline{1.23}&40.41&46.35&\bf{2.00}&47.34&58.35&1.87$\dagger$\\
  & Norm&61.40&64.38&2.04$\dagger$&38.51$\dagger$&38.84$\dagger$&2.12$\dagger$&66.62&67.53&1.30$\dagger$&40.27&\underline{47.22}&2.30$\dagger$&\underline{47.70}&\underline{58.74}&1.84$\dagger$\\
  & GC~\cite{roy2023test}&\underline{62.62}&\underline{65.38}&1.98&\underline{42.23}&\underline{42.89}&\underline{1.97}&67.03&67.83&1.25&\underline{40.94}&46.87&2.21$\dagger$&47.45&59.07&\underline{1.76}\\
  & RF~\cite{yang2023test}&61.12&64.07&1.97&40.19&40.01&2.06&67.11&67.70&1.28$\dagger$&40.18&45.98&2.28$\dagger$&47.61&58.44&1.86$\dagger$\\
  & CASP&\bf{64.23}&\bf{67.75}&\bf{1.81}&\bf{51.27}&\bf{53.15}&\bf{1.73}&\bf{69.12}&\bf{69.17}&\bf{0.96}&\bf{48.03}&\bf{50.43}&2.04&\bf{49.09}&\bf{59.11}&\bf{1.60}\\ \midrule \midrule
  \multirow{6}{*}{\makecell[c]{Early\\Fusion}}
  & Source&45.95&45.28&2.15&36.76&37.83&2.42&66.75&67.35&1.24&40.17&40.60&1.75&46.39&50.61&1.34\\
  & ST&\underline{48.80}&47.20&2.11&34.79$\dagger$&36.52$\dagger$&2.50$\dagger$&66.63$\dagger$&67.35&1.34$\dagger$&41.74&42.39&1.55&\underline{47.14}&\underline{53.68}&1.32\\
  & Norm&43.76$\dagger$&44.06$\dagger$&2.25$\dagger$&36.23$\dagger$&\underline{37.94}&2.47$\dagger$&66.98&\underline{67.54}&1.30$\dagger$&\underline{43.95}&\underline{43.40}&1.56&45.80$\dagger$&48.13$\dagger$&1.29\\
  & GC~\cite{roy2023test}&47.64&\underline{47.60}&\underline{2.10}&\underline{37.22}&37.70$\dagger$&\underline{2.29}&\underline{67.12}&67.40&\underline{1.22}&42.67&43.08&\underline{1.54}&46.77&49.12$\dagger$&\underline{1.30}\\
  & RF~\cite{yang2023test}&46.12&46.25&2.18$\dagger$&35.18$\dagger$&35.97$\dagger$&2.46$\dagger$&66.84&67.39&1.27$\dagger$&42.58&43.02&1.59&46.61&50.39$\dagger$&1.31\\
  & CASP&\bf{63.89}&\bf{66.43}&\bf{1.80}&\bf{40.12}&\bf{41.65}&\bf{2.06}&\bf{68.32}&\bf{68.90}&\bf{1.08}&\bf{46.57}&\bf{47.10}&\bf{1.44}&\bf{47.90}&\bf{57.13}&\bf{1.26}\\ \bottomrule
  \end{tabular}%
  }
  \caption{Quantitative results across five different distribution shift settings with two different backbones. For simplicity, we use MOSI, MOSEI and SIMS to represent CMU-MOSI, CMU-MOSEI and CH-SIMS. \textbf{Bold}: best results. \underline{Underline}: second best results. $\dagger$ represents that the performance decreases compared with the source model without adaptation. We report the average results using five different random seeds.}
  \label{mainresult}
\end{table*}

\section{Experiments}
\subsection{Datasets and Evaluation Metrics}
\noindent\textbf{CMU-MOSI}~\cite{7742221} is a popular dataset for multimodal (audio, text and video) sentiment analysis. It comprises 93 English YouTube videos, containing 89 distinct speakers, including 41 female and 48 male speakers. 
Each segment is manually annotated with a sentiment score ranging from strongly negative to strongly positive (-3 to +3).

\noindent\textbf{CMU-MOSEI}~\cite{bagher-zadeh-etal-2018-multimodal} is an extension of CMU-MOSI. It contains more than 65 hours of annotated video from more than 1000 speakers and 250 topics. It has a total number of 3,228 videos which are divided into 23,453 segments. 
Compared with CMU-MOSI, it covers a wider range of topics.

\noindent\textbf{CH-SIMS}~\cite{yu-etal-2020-ch} is a Chinese multimodal sentiment analysis dataset that has three modalities (audio, text and video). It has a total of 60 videos which contains 2,281 refined segments in the wild annotated with a sentiment score ranging from strongly negative to strongly positive (-1 to 1). 

For all three datasets, we use binary accuracy (ACC), F1 score (F1) and mean absolute error (MAE) as evaluation metrics.

\subsection{Baselines}
To the best of our knowledge, we are the first to introduce the TTA method to multimodal regression tasks. Therefore, previous methods can not be applied to MSA tasks due to the properties of multimodal data and the non-probabilistic model. In our experiments, we mainly compare our method with five baselines:
\textbf{Source} is the model pre-trained on the source domain data. Then it is tested on the target domain data without any adaptation strategy.
\textbf{ST} is the self-training method. We use the source model to generate pseudo labels of the target domain data. Then we use these pseudo labels to train the source model.
\textbf{Norm} is also the self-training method. Different from ST where we train the whole model, Norm only trains the normalization layers and freezes other parameters. This method~\cite{wang2020tent, roy2023test} is commonly used in TTA.
\textbf{GC} is the group contrastive strategy proposed in TTA-IQA~\cite{roy2023test}. In TTA-IAQ, the authors propose a group contrastive strategy and rank loss strategy. However, the rank loss strategy can not be applied to multimodal data due to the image augmentation strategy. Therefore, we only apply GC to our task.
\textbf{RF} is the reliable fusion strategy proposed in READ~\cite{yang2023test}. READ proposes two strategies: reliable fusion and robust adaptation. Reliable fusion strategy is a new paradigm that modulates the attention between modalities in a self-adaptive way. 
Robust adaptation is based on probabilistic models which can not be applied to regression tasks. Therefore, we use the reliable fusion strategy for comparison.

\begin{table}[]
  \centering
  \resizebox{\columnwidth}{!}{%
  \begin{tabular}{@{}l|cccccc@{}}
  \toprule
   & Case 1 & Case 2 & Case 3 & Case 4 & Case 5 & Case 6 \\ \midrule
  Epoch 0 & -2.38 & +0.83 & +2.12 & -1.21 & +1.98 & +1.68 \\
  Epoch 3 & -2.39 & +0.32 & +1.17 & -1.21 & +1.99 & +1.74 \\
  Epoch 6 & -2.40 & -0.17 & +0.85 & -1.22 & +1.95 & +1.20 \\
  Epoch 9 & -2.41 & +0.08 & +1.25 & -1.23 & +2.11 & +1.27 \\
  Epoch 12 & -2.41 & +1.02 & -0.01 & -1.23 & +2.12 & +1.01 \\
  Epoch 15 & -2.42 & +1.14 & +0.18 & -1.23 & +2.11 & +0.84 \\
  \midrule\midrule
  Stability $s$ & 0.008 & 0.46 & 0.62 & 0.004 & 0.05 & 0.22 \\
  GT & -3.0 & +2.4 & -0.8 & -1.8 & +3.0 & -0.8 \\
  Selected? & \CheckmarkBold & \XSolidBrush & \XSolidBrush & \CheckmarkBold & \XSolidBrush & \XSolidBrush \\ 
  Pseudo Label& -2.40 & - & - & -1.22 & - &-\\
  \bottomrule
  \end{tabular}%
  }
  \caption{Case study of pseudo-label generation process on MOSEI$\rightarrow$SIMS. We adapt the model for 15 epochs and set interval parameter $M=3$. The table shows the predictions of six samples. ``GT" denotes ground truth. ``Selected?" denotes whether the sample is selected for self-training. The threshold is 0.012 as shown in Figure~\ref{pse} when $\lambda=95$.}
  \label{gene}
\end{table}

\subsection{Implementation Details}
\noindent\textbf{Raw Feature Extraction.} For text modality, we use pre-trained BERT~\cite{devlin2019bert} to obtain word embeddings. We use BERT-base for CMU-MOSI and CMU-MOSEI and Chinese BERT-base for CH-SIMS. Each word is represented as a 768-dimensional vector. For audio modality, we use LibROSA~\cite{McFee2015librosaAA} to extract features. For video modality, we extract face features using OpenFace 2.0~\cite{8373812} toolkit.

\noindent\textbf{Source Domain and Target Domain.} We denote source domain to target domain as $A\rightarrow B$ where $A$ represents the source domain and $B$ is the target domain. We validate CASP across five distribution shift settings: CMU-MOSEI$\rightarrow$CH-SIMS, CMU-MOSI$\rightarrow$CH-SIMS, CMU-MOSI$\rightarrow$CMU-MOSEI, CH-SIMS$\rightarrow$CMU-MOSI and CH-SIMS$\rightarrow$CMU-MOSEI. We do not use CMU-MOSEI$\rightarrow$CMU-MOSI because CMU-MOSEI is an extension of CMU-MOSI and covers a wider range of topics compared with CMU-MOSI.

\noindent\textbf{Backbones.} To demonstrate the generalization ability of our method, we use two different backbones: the feature-level fusion (early fusion) method and the decision-level fusion (late fusion) method. Specifically, we use the transformer encoder~\cite{vaswani2017attention} as the backbone. For fairness, we use the same backbone for all the methods.

\noindent\textbf{Training Details.} For source domain pre-training and contrastive adaptation, we use the AdamW optimizer with a learning rate of $1e-3$. We adapt the model for 15 epochs and the interval hyperparameter $M$ is set to 3. For stable pseudo-label generation, we set the threshold hyperparameter $\lambda$ as 95. For self-training using stable pseudo labels, we use the AdamW optimizer with a learning rate of $5e-4$ and train the model for 5 epochs. The batch size of all the experiments is 24. Besides, we use gradient clipping and set the threshold as 0.8. We also use a step scheduler with a step size of 10 and decay rate $\gamma=0.1$. To avoid randomness, we train the model five times using five different random seeds and report the average results.

\begin{figure}
  \centering
  \includegraphics[width=0.78\linewidth]{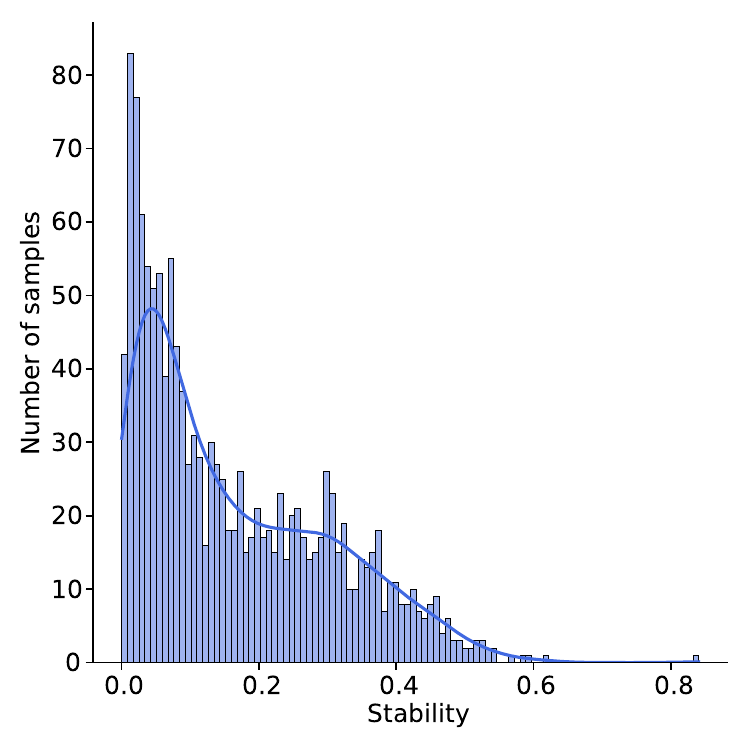}
  \caption{The distribution of stability $s$ on MOSEI$\rightarrow$SIMS.}
  \label{pse}
  \vskip -0.1in
\end{figure}

\subsection{Main Results}
We present our quantitative results across five different distribution shift settings with two different backbones in Table~\ref{mainresult}. Compared with the source model without any adaptation strategy, CASP brings significant and consistent performance improvements across all settings and with different backbones. ST, Norm, GC and RF are four different methods to mitigate the distribution shifts between the source domain and the target domain. However, we observe that all these four methods have performance degradation on some metrics across some distribution shifts, which are marked with $\dagger$ in the table. Only CASP can bring consistent performance improvements.

Besides, the source model performs very poorly on MOSEI$\rightarrow$SIMS, MOSI$\rightarrow$SIMS, SIMS$\rightarrow$MOSI and SIMS$\rightarrow$MOSEI because SIMS is a Chinese multimodal sentiment analysis dataset while MOSI and MOSEI are English multimodal datasets. Therefore, SIMS has a huge distribution shift from MOSI and MOSEI. The accuracies across these settings are below 50\%. All the baselines bring limited improvements across these settings while CASP brings significant improvements. On MOSEI$\rightarrow$SIMS, CASP improves the accuracy of the early fusion backbone by nearly 20\% while the second best method ST improves the accuracy by just around 3\%. On MOSI$\rightarrow$SIMS, CASP improves the accuracy of the late fusion backbone by around 11\% while the second best method GC improves the accuracy by just around 3\%. Only on SIMS$\rightarrow$MOSI, ST performs better on MAE with 0.04 higher than that of CASP. These results fully demonstrate the superiority and versatility of CASP.

\subsection{Stable Pseudo-label Generation}
In Figure~\ref{pse}, we present the distribution of stability $s$ on MOSEI$\rightarrow$SIMS. As shown in the figure, most of the samples do not change drastically during the contrastive adaptation process. When we use the most stable 5\% of the samples as pseudo labels (i.e. the hyperparameter $\lambda=95$), the stability threshold is 0.012. We present the changes of six samples during the contrastive adaptation process in Table~\ref{gene}. From Case 2, Case 3 and Case 6, we can find that these three cases have high stability $s$ (\textit{i.e.} not stable) and their predictions differ greatly from the ground truths. This demonstrates that high stability means low-confident labels, indicating the effectiveness of our method in choosing high-confident pseudo labels. Comparing Case 1 and Case 5, we can find that although Case 5 also has low stability $s$, the difference between the prediction and the ground truth of Case 5 is bigger than that of Case 1, indicating that lower stability means higher confidence. Moreover, from epoch 0 (\textit{i.e.} Source model) to epoch 15, it is easy to find that the predictions change towards the ground truths. For example, at epoch 0, the prediction of Case 1 is -2.38, and at epoch 15, the prediction is -2.42, which is closer to the ground truth -3.0. This also demonstrates the effectiveness of our contrastive adaptation.

\begin{table}
  \centering
  \resizebox{0.97\columnwidth}{!}{%
  \begin{tabular}{@{}c|cc|ccccc@{}}
  \toprule
  Backbone                      & CA & SPL & $T_1$  & $T_2$ & $T_3$ & $T_4$ & $T_5$ \\ \midrule
  \multirow{3}{*}{\makecell[c]{Late\\Fusion}}  
  &\XSolidBrush&\XSolidBrush&60.96&39.17&66.57&40.12&47.14\\
  &\CheckmarkBold&\XSolidBrush&62.36&49.45&67.74&41.15&48.33\\
  &\CheckmarkBold&\CheckmarkBold&64.23&51.27&69.12&48.03&49.09\\ \midrule \midrule
  \multirow{3}{*}{\makecell[c]{Early\\Fusion}} 
  &\XSolidBrush&\XSolidBrush&45.95&36.76&66.75&40.17&46.39\\
  &\CheckmarkBold&\XSolidBrush&61.62&38.95&67.76&44.10&47.23\\
  &\CheckmarkBold&\CheckmarkBold&63.89&40.12&68.32&46.57&47.90\\ \bottomrule
  \end{tabular}%
  }
  \caption{Quantitative results of contributions of the contrastive adaptation (CA) and stable pseudo-label generation (SPL). $T_1,T_2,T_3,T_4,T_5$ represents the distribution shift: MOSEI$\rightarrow$SIMS, MOSI$\rightarrow$SIMS, MOSI$\rightarrow$MOSEI, SIMS$\rightarrow$MOSI and SIMS$\rightarrow$MOSEI, respectively. We only report the accuracy in the table.}
  \label{ab1}
\end{table}

\begin{table}[]
  \centering
  \resizebox{0.93\columnwidth}{!}{%
  \begin{tabular}{@{}c|c|ccccc@{}}
  \toprule
  Backbone& $n$ & $T_1$ & $T_2$ & $T_3$ & $T_4$ & $T_5$ \\ \midrule
  \multirow{3}{*}{\makecell[c]{Late\\Fusion}} 
  &none&60.96&39.17&66.57&40.12&47.14  \\
  &1&62.36&49.45&67.74&41.15&48.33 \\
  &2&61.11 & 44.67& 66.94 & 40.29 & 47.60 \\ \midrule \midrule
  \multirow{3}{*}{\makecell[c]{Early\\Fusion}} 
  &none&45.95&36.76&66.75&40.17&46.39 \\
  &1&61.62&38.95&67.76&44.10&47.23\\
  &2&53.12&36.94&66.81&41.57&46.80\\ \bottomrule
  \end{tabular}%
  }
  \caption{The impact of the number of dropped modalities. The table shows the accuracy of the model. $n$ represents the number of dropped modalities. $T_1,T_2,T_3,T_4$ and $T_5$ have the same meaning as Table~\ref{ab1}.}
  \label{ab4}
\end{table}

\subsection{Ablation Study}
In this subsection, we conduct a series of ablation experiments for a better understanding of CASP. Concretely, we will analyze the contribution of contrastive adaptation and stable pseudo-label generation. Besides, we will explore the impact of the number of dropped modalities, the quality of the selected pseudo labels, the impact of the interval hyperparameter $M$ and the stability threshold $\lambda$ on the performance of the model.

\noindent\textbf{Contributions of contrastive adaptation and stable pseudo-label generation.} Table~\ref{ab1} presents the accuracy of the ablation experiments. We observe that both the contrastive adaptation strategy and stable pseudo-label generation improve the performance of the model. Particularly, compared to the stable pseudo-label generation strategy, the contrastive adaptation strategy has more potential to improve the performance of the model. Across five different distribution shifts, self-training with stable pseudo labels improves the accuracy by around 1\%-7\% while the contrastive adaptation strategy can improve the accuracy by up to around 10\%-15\%. Furthermore, we observe that contrastive adaptation helps more when the accuracy of the source model is low and helps less when the accuracy of the source model is high. In contrast, the stable pseudo-label strategy brings relatively steady and consistent improvements to the performance of the model.

\begin{table}[h!]
  \centering
  \resizebox{\columnwidth}{!}{%
  \begin{tabular}{@{}c|c|ccccc@{}}
  \toprule
  Backbone                      & Method & $T_1$ & $T_2$ & $T_3$ & $T_4$ & $T_5$ \\ \midrule
  \multirow{4}{*}{\makecell[c]{Late\\Fusion}} 
  &SPL ($M=1$)& 40.25& 37.25 & 47.08 & 44.98 &52.01  \\
  &SPL ($M=2$)&  41.07& 38.55 & 47.60 &45.34  & 55.17 \\
  &SPL ($M=3$)& 42.11 & 38.99 & 48.86 & 45.22 & 57.81 \\
  &SPL ($M=4$)&42.23 & 38.89 & 49.55 & 46.80 & 59.21 \\ \midrule \midrule
  \multirow{4}{*}{\makecell[c]{Early\\Fusion}} 
  &SPL ($M=1$)& 46.11 & 45.20 & 46.14 & 52.01 & 51.12 \\
  &SPL ($M=2$)& 46.85 & 45.56 & 46.55 & 52.17 & 51.02 \\
  &SPL ($M=3$)& 46.61 & 46.06 & 47.69 & 54.70 & 52.48 \\
  &SPL ($M=4$)& 46.80 & 46.23 & 47.71 & 54.57 & 53.96 \\ \bottomrule
  \end{tabular}%
  }
  \caption{The impact of the interval hyperparameter $M$ on the quality of the stable pseudo labels. The table reports the rates of MAE decline compared to the pseudo labels obtained directly using the source model. In all the experiments, we fix the adaptation epoch $E=20$. $T_1,T_2,T_3,T_4$ and $T_5$ have the same meaning as Table~\ref{ab1}.}
  \label{ab2}
\end{table}

\begin{figure}
  \centering
  \includegraphics[width=1\linewidth]{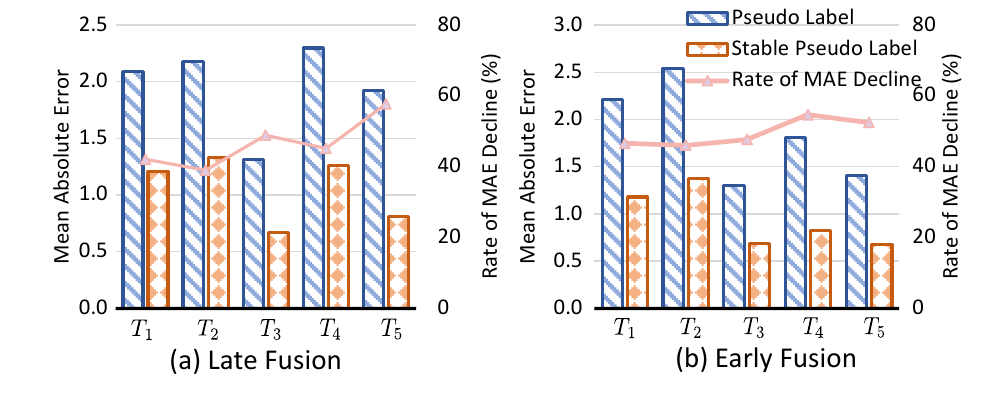}
    \caption{Effectiveness of the stable pseudo-label generation strategy across five different distribution shift settings and with two different backbones. $T_1,T_2,T_3,T_4$ and $T_5$ have the same meaning as Table~\ref{ab1}.}
    \label{spl}
\end{figure}

\noindent\textbf{The number of dropped modalities.} In the contrastive adaptation strategy, we randomly drop a modality as a way of augmentation. To explore the impact of the number of dropped modalities, we conduct experiments and present the results in Table~\ref{ab4}. Both $n=1$ and $n=2$ bring consistent improvements to the performance of the model. Furthermore, we observe that the performance of the model only dropping one modality is better than that of the model dropping two modalities. In our datasets, there are only three modalities. Therefore, dropping two modalities may lose a lot of information and enforce strong consistency, which may hinder the performance of the model compared with only dropping one modality. If a dataset has more modalities such as five modalities, dropping two modalities may be a better choice.

\noindent\textbf{Quality of the stable pseudo labels.} To demonstrate that the quality of our stable pseudo labels is much better than that of pseudo labels obtained directly using the source model, we calculate the mean absolute error between the pseudo labels and the ground truths and present our results in Figure~\ref{spl}. From the figure, we observe that the MAE of stable pseudo labels is much lower than that of pseudo labels obtained using the source model. The rates of MAE decline across five different distribution shift settings and two different backbones are all around 40\%-60\%, demonstrating the effectiveness of our stable pseudo-label generation strategy.

\noindent\textbf{The impact of the interval hyperparameter $M$.} To explore the impact of the interval hyperparameter $M$ on the quality of stable pseudo labels, we select $M=1,2,3,4$ and generate stable pseudo labels to calculate the rates of MAE decline compared to the pseudo labels generated by the source model. We report our results in Table~\ref{ab2}. We observe that the quality of the stable pseudo labels increases as $M$ increases. Intuitively, when the interval $M$ is large, the difference between the two checkpoints is large. This is beneficial to the selection of high-confident labels because our stable pseudo-label generation strategy calculates the average value of the difference between two consecutive checkpoints and selects the labels whose values are lower than a threshold. A large $M$ means large differences and thus would help to select these stable labels. However, to include relatively more checkpoints, a large $M$ requires more training epochs $E$, which could increase the time of the adaptation process. To get a balance between the adaptation time and the performance, $M=2,3$ would be an opportune value.

\begin{table}[]
  \centering
  \resizebox{0.9\columnwidth}{!}{%
  \begin{tabular}{@{}c|c|ccccc@{}}
  \toprule
  Backbone& $\lambda$ & $T_1$ & $T_2$ & $T_3$ & $T_4$ & $T_5$ \\ \midrule
  \multirow{3}{*}{\makecell[c]{Late\\Fusion}} 
  &50&  12.01 & 10.17 & 13.22 & 13.67  & 19.90 \\
  &75& 30.66 & 21.85 & 27.68 & 31.02 & 41.10 \\
  &95& 42.11 & 38.99 & 48.86 & 45.22 & 57.81 \\ \midrule \midrule
  \multirow{3}{*}{\makecell[c]{Early\\Fusion}} 
  &50& 13.17 & 12.01 & 13.22 & 19.34 & 17.25 \\
  &75& 26.12 & 22.20 & 23.41 & 31.46 & 30.14 \\
  &95& 46.61 & 46.06 & 47.69 & 54.70 & 52.48 \\ \bottomrule
  \end{tabular}%
  }
  \caption{The impact of the threshold $\lambda$ on the quality of the stable pseudo labels. The table reports the rates of MAE decline compared to the pseudo labels generated using the source model. $T_1,T_2,T_3,T_4$ and $T_5$ have the same meaning as Table~\ref{ab1}.}
  \label{ab3}
\end{table}

\noindent\textbf{The impact of the threshold $\lambda$.} We select three different threshold $\lambda$ and present our ablation results in Table~\ref{ab3}. The results demonstrate our stable pseudo-label generation strategy can generate high-confident pseudo labels whatever $\lambda$ is. With the increase of $\lambda$, the quality of the pseudo labels also increases. However, with the increase of $\lambda$, the number of samples for self-training decreases. Therefore, if the test set has many samples, we are expected to choose a large $\lambda$ while if the test set does not have many samples, we can decrease the value of $\lambda$.

\section{Conclusion}
In this paper, we focus on the test-time adaptation for multimodal sentiment analysis. Due to the reason that multimodal sentiment analysis is a multimodal regression task, existing methods can not be applied. To address the distribution shifts for multimodal sentiment analysis, we propose contrastive adaptation and stable pseudo-label generation (CASP) strategies. CASP has two stages: contrastive adaptation to enforce consistency and self-training with stable pseudo labels to minimize empirical risk. We conduct extensive experiments across various distribution shift settings and with different backbones. The results demonstrate the superiority and versatility of CASP. Ablation experiments are then conducted to validate the main components of CASP.

\bibliography{aaai25}

\end{document}